\documentclass[11pt]{article}

\usepackage[margin=1in]{geometry}
\usepackage{times}
\usepackage[T1]{fontenc}
\usepackage[utf8]{inputenc}
\usepackage{amsmath,amssymb}
\usepackage{graphicx}
\usepackage{booktabs}
\usepackage{natbib}
\usepackage{hyperref}
\usepackage{xcolor}
\usepackage{caption}

\hypersetup{
  colorlinks=true,
  linkcolor=blue,
  citecolor=blue,
  urlcolor=blue
}

\title{Attribution Gaps in Zero-Training LLM+OVOD Pipelines: \\
A Fine-Grained Analysis of the CAAP--SNAP Discrepancy}

\author{
  Yu-Feng Yen \\
  Independent Researcher \\
  Taiwan \\
  \texttt{ccuhollis@alum.ccu.edu.tw}
}

\date{}

\begin{document}

\maketitle

\begin{abstract}
Zero-training LLM+open-vocabulary-detector (OVOD) pipelines such as LAOD report class-agnostic localization (CAAP) and semantic naming (SNAP) separately; with CLIP text similarity SNAP is far above CAAP (0.46 against 0.21 on COCO-Val). Much of this gap comes from CLIP's high similarity floor at low thresholds: with shuffled labels the CLIP score is still 0.33, and with Sentence-BERT SNAP is 0.22, close to CAAP. We then ask what the localization failures that CAAP counts are, on 27{,}273 detections from 5{,}000 images. The IoU-0.5 hit rate against COCO boxes varies more with vocabulary than with size: 80.9\% of detections with labels inside COCO's 80 categories reach IoU 0.5, against 31.0\% outside, and the effect reproduces with another detector and another LLM. Of the 12{,}139 failures, only 14.3\% are loose boxes on the named object. Audits with the label visible (n = 200 and 96) estimate that 73\% of all failures are real objects COCO does not annotate (63\% if stuff regions such as sky and wall are set aside). A label-free audit by two other annotators, who count stuff regions as not valid, gives 77\% and 66\%, against 63\% for the label-visible audits on the same basis; one lower bound is below our pre-registered 60\% threshold, so we state the valid share as a range, about two thirds to just over three quarters, and not as 73\%. LVIS re-scoring gives a lower bound of 39.6\%; COCO-ReM shows that missing instances of COCO's 80 classes explain only 6.3\%. Much of what CAAP scores as localization failure reflects closed-category annotation; because the pipeline recalls only 38.7\% of COCO's own boxes, how far CAAP itself would rise is left open.
\end{abstract}

\section{Introduction}

Open-world detection increasingly relies on zero-training pipelines that pair a large language model (LLM), which names the objects in a scene, with an open-vocabulary detector (OVOD) that localizes whatever the LLM names. LAOD (\citealp{laod2025}) is a recent instance. It reports two scores: CAAP (class-agnostic average precision: is the box in the right place, whatever the label?) and SNAP (semantic naming average precision: is the label semantically close to a ground-truth category?). In its aggregate scores SNAP is well above CAAP; the paper reports this and does not investigate the cause.

This is an analytical study; we propose no new model or metric. We ask whether the gap belongs to the pipeline or to the text encoder behind SNAP, and then what the failures that CAAP counts are, and how much of them is a capability limit rather than an artifact of how COCO is annotated. Our claims are:

\begin{enumerate}\setlength\itemsep{2pt}
\item \textbf{Much of the CAAP--SNAP gap is an encoder effect.} We reproduce the gap as LAOD defines it (CAAP 0.21, SNAP 0.46 on the full COCO-Val split) and show that a label-shuffled control scores 0.33 under CLIP, and that with Sentence-BERT SNAP is 0.22 (\S{}3.1). LAOD's rewritten repository (2026-10) documents the same encoder problem.
\item \textbf{Vocabulary novelty predicts failure to match COCO boxes more than object size does}, and within novel vocabulary the drop concentrates in labels that name a different object than the annotated one (\S{}3.2, \S{}3.3). The effect reproduces with a different detector and a different LLM (\S{}3.5).
\item \textbf{Most CAAP failures are valid detections of objects COCO does not annotate.} We split the 12{,}139 IoU-0.5 failures into no overlap with any ground-truth box (32.5\%), overlap only with a differently-labelled box (53.3\%) and loose boxes on the named object (14.3\%). Audits with the label visible estimate 73\% of failures to be valid but unannotated (63\% with stuff regions not counted); a label-free audit by two other annotators, who do not count stuff regions, gives 77\% and 66\%, and one lower bound is below our pre-registered 60\% threshold, so we state the share as about two thirds to just over three quarters (\S{}3.4). Missing instances of COCO's 80 classes explain only about 6\% of the failures (\S{}3.6).
\end{enumerate}

\subsection{Related work}

LAOD prompts an LLM for scene-specific object names and passes them to an open-vocabulary detector; we use its Gemma-3 and YOLO-World components and additionally test Grounding DINO (\citealp{liu2024groundingdino}) and Qwen2.5-VL. TIDE (\citealp{bolya2020tide}) decomposes detection errors by type; we apply the idea to CAAP/SNAP. COCO-ReM (\citealp{singh2024cocorem}), COCONut (\citealp{deng2024coconut}) and LVIS (\citealp{gupta2019lvis}) document that COCO annotation is incomplete or non-exhaustive. NoOVD (\citealp{zhang2026noovd} per its arXiv entry; arXiv:2603.21069) argues that objects a model discovers but COCO does not label count as false positives, without measuring it; B-FOR (\citealp{batool2026bfor} per its arXiv entry; arXiv:2608.29232) evaluates class-agnostic detectors with recall for the same reason and, on the top-1 prediction of 100 COCO val images, finds 48\% and 42\% valid but unlabeled objects. We analyse a zero-training LLM+OVOD pipeline at the scale of 27{,}273 detections, with the failures split into three audited groups. Our literature searches were four arXiv queries on 2026-10-09.

\subsection{Corrections relative to arXiv v1}

This version corrects arXiv v1 (2609.32567). (1) v1's corpus CAAP (0.38; 0.25 on the pilot) was inflated by a bug that matched pooled detections against the boxes of all images; with per-image matching it is 0.21 (pilot 0.20), within 0.01--0.03 of LAOD's values in each interval. (2) That tiny objects score higher than large ones was a selection artifact. (3) The ``right object, loose box'' mechanism rested on a similarity criterion that 99.4\% of detections satisfy. (4) The three-way vocabulary table came from an undocumented judge and is replaced by a documented two-way contrast. (5) v1's conclusion that ``a large share of the apparent CAAP--SNAP gap traces back to closed-category annotation limits'' is withdrawn: the audits support it for CAAP failures, not for the gap, which \S{}3.1 shows is in large part an encoder effect. Also withdrawn: that small and occluded objects are localized better (v1 tested size only); the 78--88\% share of real unannotated objects (for the same no-overlap group 75.5\% [69, 81], and 73\% of all failures); that the LLM's naming is rarely the problem; and that the asymmetry's persistence at 24 times the sample size rules out a small-sample effect (the pilot is a subset of the same run).

\section{Method}

\textbf{Pipeline.} We use LAOD's pipeline: an LLM (Gemma-3 4B; Qwen2.5-VL 7B as a check) names the singular objects in an image with LAOD's prompt, and the label set goes to an open-vocabulary detector (YOLO-World X; Grounding DINO as a check). Nothing is fine-tuned, and the LLM is sampled without a seed. LAOD's public repository had no batch evaluation code when we ran our experiments (it was rewritten on 2026-10-06), so we implemented a COCO loader, CAAP and a CLIP-based SNAP. The code and the predictions are released (Appendix).

\textbf{Metrics.} CAAP is IoU-based average precision with the class ignored, averaged over IoU 0.50--0.95 (LO, MI and HI are LAOD's low, medium and high IoU intervals), matching each detection to a ground-truth box of its own image one-to-one. SNAP follows LAOD's Eqs. 8--12: a label is a true positive if its CLIP text-embedding cosine similarity to some unmatched ground-truth label of the image is at least $\tau$, and SNAP is the mean over $\tau$ = 0.6, 0.7, 0.8, 0.9 of the area under the resulting precision--recall curve. CLIP text similarity has a high baseline (banana/airplane scores 0.73): 99.4\% of detections have similarity $\geq$ 0.7 with some ground-truth label of their image, so analyses that need a discriminative criterion use exact class-name matching.

\begin{center}\footnotesize\setlength{\tabcolsep}{3pt}
\begin{tabular}{p{0.200\linewidth}p{0.370\linewidth}p{0.370\linewidth}}
\toprule
 & CAAP, LO / MI / HI & SNAP \\
\midrule
ours (5{,}000 images) & 0.268 / 0.231 / 0.113 & mean 0.456 ($\tau$ = 0.6: 0.559, 0.9: 0.251) \\
LAOD (reported) & 0.25 / 0.22 / 0.08 & LO 0.54 / MI 0.52 / HI 0.19 \\
ours over the thresholds LAOD's code produced & 0.264 / 0.231 / 0.085 & LO 0.559 / MI 0.529 / HI 0.203 \\
\bottomrule
\end{tabular}
\end{center}

Our CAAP is slightly above LAOD's in every interval. The difference at HI (0.113 against 0.08) is not explained by AP interpolation, class-agnostic NMS or crowd boxes. It is explained by the thresholds LAOD's code produced: its rewritten repository (commit 0d57f25; the arXiv paper is still v1) documents that the paper's intervals were built with \texttt{np.\allowbreak{}arange} calls that emit four values at floating-point precision, so LO also contains 0.65 and HI contains 0.85, 0.90, 0.95 and 1.00, where almost no detection reaches IoU 1.00. Over those thresholds our CAAP is within 0.014 of LAOD's (computed from stored per-threshold values, without running LAOD's code). Elsewhere we use the disjoint intervals (LO 0.50--0.60, MI 0.65--0.80, HI 0.85--0.95).

\textbf{Data.} We evaluate on the full COCO-Val split (5{,}000 images, 27{,}273 detections). COCO annotations are CC BY 4.0; images are not redistributed. The LLM judges of \S{}3.3 and the naming check of \S{}3.4 were called through OpenRouter with text only, never images, with the training-on-requests setting off.

\section{Findings}

\subsection{Much of the CAAP--SNAP gap is an encoder effect}

On the full split the pipeline gives corpus CAAP 0.207 and SNAP 0.456 with CLIP (image-level r = 0.758), a gap of 0.249; interval by interval our 0.268 against 0.559 matches LAOD's 0.25 against 0.54. The 200-image pilot (CAAP 0.203, SNAP 0.438) is a subset of the same run and is not an independent replication.

LAOD's rewritten repository raises the same concern about CLIP-based SNAP (a median CLIP text cosine of 0.786 between unrelated COCO labels, so low thresholds admit every pair) and replaces CLIP by Sentence-BERT. We tested this directly by recomputing SNAP on our detections with a copy in which the predicted labels are shuffled across detections. With CLIP the shuffled copy scores 0.558--0.559 at every $\tau$ from 0.50 to 0.65, and 0.330 averaged over our four thresholds, against 0.456 real; it falls away only at higher thresholds (0.251 against 0.004 at $\tau$ = 0.9). With Sentence-BERT (all-MiniLM-L6-v2) the real labels give 0.224 over the four thresholds (0.230 over ten) against 0.003 shuffled, close to our CAAP of 0.207; interval by interval SNAP exceeds CAAP by only 0.004 (LO), 0.001 (MI) and 0.072 (HI). Much of the gap measured with CLIP is therefore produced by CLIP at low thresholds, and we claim no particular size for the true difference between naming and localization.

This comparison has limits. The two scores average over different grids, so ``close'' depends on where one reads it. Sentence-BERT scores are probably pushed down by COCO's closed vocabulary (a valid synonym or a name for an unannotated object scores low against annotated labels), so they are not a naming-accuracy estimate either. The shuffle uses one seed. Our ten-threshold value (0.230) is below LAOD's 0.270 for the same model pair, on a different split and detection cap; we did not reproduce theirs.

\subsection{Failure to match COCO boxes tracks vocabulary novelty more than object size}

A detection is an IoU-0.5 hit if its IoU with some ground-truth box of its image is at least 0.5, and otherwise an IoU-0.5 failure, zero-overlap detections included; we avoid the term ``false positive'' because the audits below show most failures are correct detections.

\textbf{Size.} Bucketing by the size of the matched ground-truth box suggests that tiny objects score higher than large ones (71.0\% vs. 49.9\%), but this is a selection artifact: a detection matches a tiny box only if it lies almost on top of it. Bucketing by detection-box size gives the expected direction, and splitting by vocabulary category (Fig. 1) the hit rate rises with size for native labels (71.7\% to 94.7\%) and for out-of-vocabulary labels (16.8\% to 47.5\%).

\begin{figure}[t]
\centering
\includegraphics[width=0.7\textwidth]{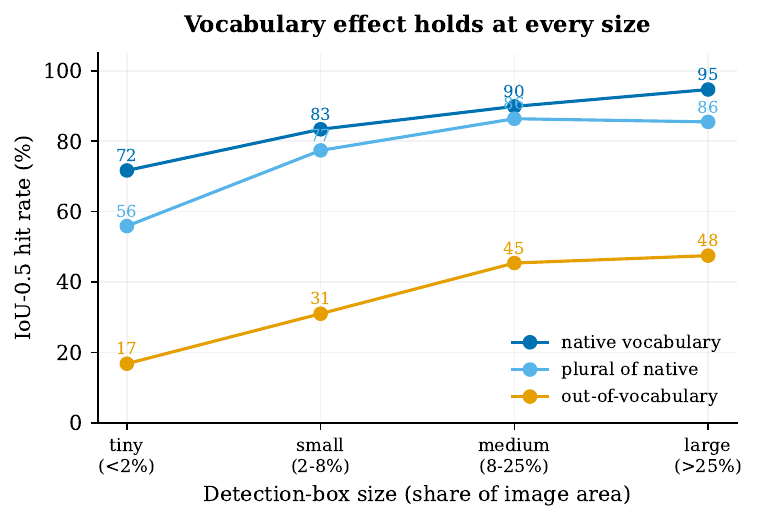}
\caption{Hit rate by detection-box size, split by vocabulary category.}
\label{fig:1}
\end{figure}

\textbf{Vocabulary.} Detections whose label matches a native COCO-80 category reach an 80.9\% hit rate (n = 11{,}557), against 31.0\% (n = 13{,}156) for out-of-vocabulary labels, a gap of 49.9 points (95\% interval 48.1--51.8 resampling images). Out-of-vocabulary labels are 48.2\% of detections, so this is not an edge case (Fig. 2). The gap stays at 44--55 points in every size bin, larger than the 23--31 points that size spans within a category.

\begin{figure}[t]
\centering
\includegraphics[width=0.7\textwidth]{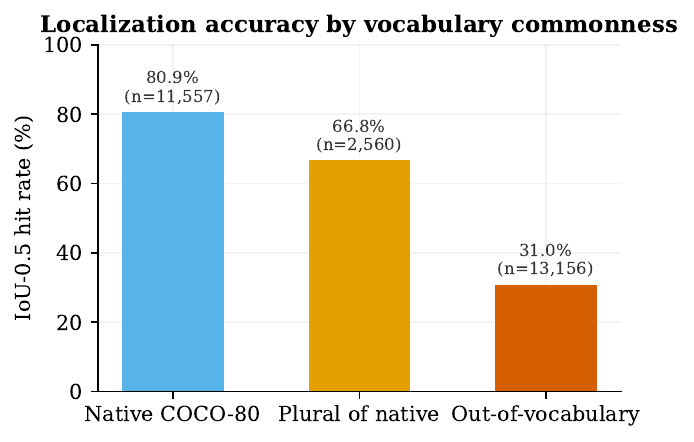}
\caption{Hit rate by vocabulary category (native, plural of native, out-of-vocabulary). Out-of-vocabulary labels fall to 31.0\% from 80.9\%.}
\label{fig:2}
\end{figure}

\textbf{Failure modes.} With exact class-name matching (plural-normalised, person descriptors, a short synonym list), the 12{,}139 failures split into three groups (Fig. 3): 3{,}940 (32.5\%) with no overlap with any ground-truth box (IoU < 0.02); 6{,}468 (53.3\%) overlapping only a box of a different class, dominated by pairs such as helmet$\to$person (371), hat$\to$person (262) and shoes$\to$person (183); and 1{,}731 (14.3\%) loose boxes on the named object (a name-matching count that can miss loose boxes carrying other labels and was not verified manually). Imprecise box regression is thus a minority mechanism, and what remains is a mismatch between what the LLM names and what COCO annotates.

\begin{figure}[t]
\centering
\includegraphics[width=0.7\textwidth]{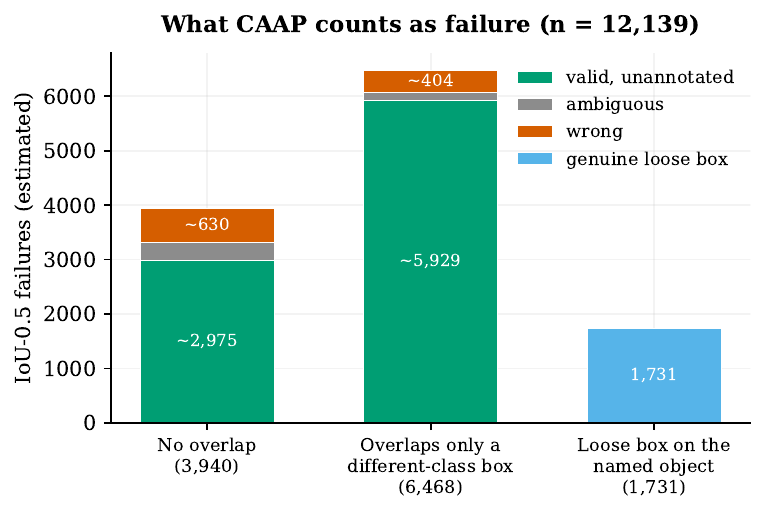}
\caption{The 12{,}139 IoU-0.5 failures by group; the audits split the first two groups into valid-but-unannotated, ambiguous and wrong, extrapolated from n = 200 and n = 96, so the counts in the first two bars are estimates.}
\label{fig:3}
\end{figure}

\subsection{Labels that name a different object than the annotated one}

We classified the 771 out-of-vocabulary labels with at least one ground-truth match by their relation to the COCO category their boxes most often overlap (\emph{synonym}, \emph{finer-grained}, \emph{out-of-ontology}, the last being a different object, part or accessory such as a helmet, shelf or plate), with three LLM-judge runs (two prompts of DeepSeek-V4-Flash and Mistral Large 4; temperature 0, prompts fixed in advance) and a human labelling of 50 labels. The three-way split is not stable enough to report: the main judge missed the pre-fixed agreement threshold ($\kappa$ = 0.51 against the human labels, threshold 0.6). Collapsed post hoc to \emph{same object} (synonym or finer-grained) versus \emph{different object} (out-of-ontology), a robustness view and not the pre-registered analysis, the judges reach $\kappa$ = 0.65, 0.72 and 0.80 against the human labels (on the 37 labels whose words do not appear in the prompt).

\begin{center}\footnotesize\setlength{\tabcolsep}{3pt}
\begin{tabular}{p{0.200\linewidth}p{0.370\linewidth}p{0.370\linewidth}}
\toprule
Judge & same-object labels: hit rate & different-object labels: hit rate \\
\midrule
DeepSeek-V4-Flash, prompt v1 & \textbf{82.1\%} & \textbf{12.5\%} \\
DeepSeek-V4-Flash, prompt v2 & \textbf{86.2\%} & \textbf{13.6\%} \\
Mistral Large 4, prompt v1 & \textbf{86.6\%} & \textbf{13.7\%} \\
\bottomrule
\end{tabular}
\end{center}

\emph{Hit rates over detections that overlap some ground-truth box (Fig. 4).}

\begin{figure}[t]
\centering
\includegraphics[width=0.7\textwidth]{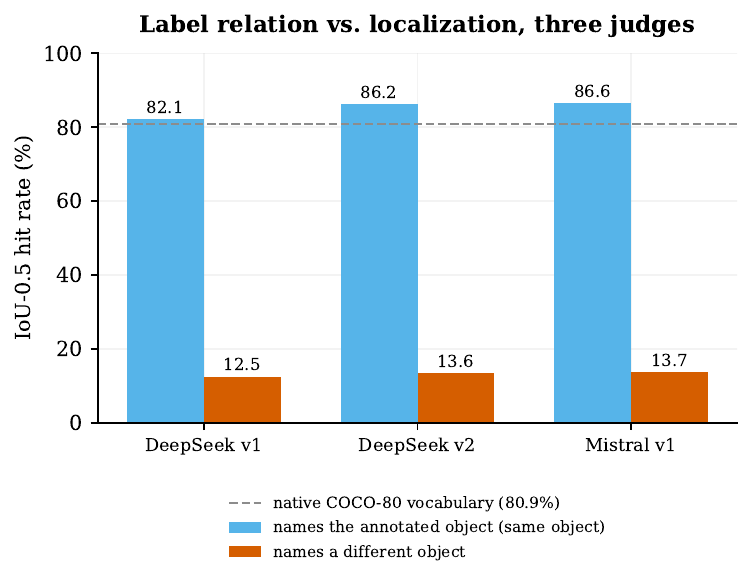}
\caption{Hit rate of same-object versus different-object labels under three LLM judges, over detections that overlap a ground-truth box. The dashed line is the native rate over all detections (80.9\%; 82.7\% on the bars' denominator).}
\label{fig:4}
\end{figure}

Labels naming a different object are localized at about 13\%, against 82--87\% for labels naming the annotated object itself (native labels: 82.7\% on the same denominator). Because both a label's class and the hit outcome depend on overlap with COCO boxes, the contrast is partly circular. A text-only variant removes the overlap: two judges map each label to one of COCO's 80 categories from the word alone, or to \emph{none}. Labels given a category reach 86.9\% and 83.7\%, labels given \emph{none} 15.8\% and 14.9\%, gaps of 71.1 and 68.7 points, against 70--73 with the overlap-based reference. This variant has no human reference and the outcome still depends on COCO boxes, so it shows that failures concentrate in labels for things COCO does not annotate, not that those boxes are correct; \S{}3.4 tests that.

\subsection{Audits: are the failures real?}

\textbf{Label visible.} We drew 200 cases at random from the no-overlap group and 96 from the group overlapping only a differently-labelled box, overlaid the box and all COCO boxes, and classified each as \emph{valid but unannotated} (the box encloses a real object the label names, and COCO has no box for it), \emph{ambiguous} or \emph{wrong}. The labels are one author's; a second, anonymous annotator labelled all 296 cases on the same form.

\begin{center}\footnotesize\setlength{\tabcolsep}{3pt}
\begin{tabular}{p{0.200\linewidth}p{0.240\linewidth}p{0.240\linewidth}p{0.240\linewidth}}
\toprule
Group (n) & valid but unannotated & wrong & ambiguous \\
\midrule
no overlap (200), author & 151 (75.5\%) [69.1, 80.9] & 32 (16.0\%) & 17 (8.5\%) \\
different-class overlap (96), author & 88 (91.7\%) [84.4, 95.7] & 6 (6.3\%) & 2 (2.1\%) \\
\bottomrule
\end{tabular}
\end{center}

\emph{Wilson 95\% intervals; Fig. 5 plots the first row. Examples: a trash can at confidence 0.90, gloves at 0.93.}

\begin{figure}[t]
\centering
\includegraphics[width=0.7\textwidth]{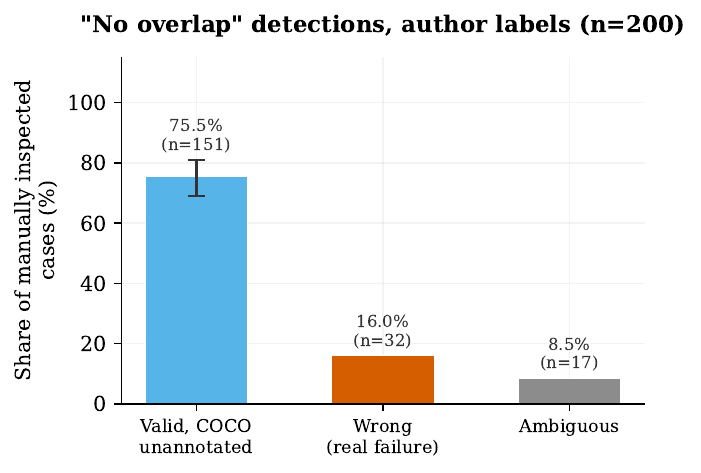}
\caption{Manual classification of the no-overlap subset (n = 200); the error bar is the Wilson 95\% CI for the valid-but-unannotated share.}
\label{fig:5}
\end{figure}

Combining the two audits, an estimated 73\% of IoU-0.5 failures (95\% interval 69--76\%) are valid detections COCO does not annotate, 14.3\% are loose boxes, about 9\% clear failures and about 4\% ambiguous. The second annotator gives 74.1\% [69.4, 77.1]; 70.0\% if both must agree and 77.5\% if either does; agreement is moderate ($\kappa$ = 0.60 for valid versus not). Some valid detections name a stuff label (sky, wall, field, beach, sea: 5 of the 151 in the first audit, 18 of the 88 in the second); setting these aside lowers the valid share to 62.5\% [56.9, 67.1]. Counting the valid detections as true positives would raise detection-level precision from 55.5\% to about 88\%. This is precision, not AP, because adding the unannotated objects also enlarges the ground-truth set. Both annotators saw each box with its label, so anchoring is not excluded.

\textbf{Label free.} We therefore ran a second audit with the protocol fixed in writing before any annotator saw an item. It has 200 boxes in random order: 100 test boxes drawn from the two audits (65 and 35; from the 264 of the 296 audited cases that could be traced to the detection table), 50 positive controls (CAAP hits) and 50 negative controls (hit boxes moved off every COCO box or shrunk to about 36\% of their area). Each item shows the image and an enlarged crop with the box in red; the label, class and COCO boxes are hidden. Two annotators, L1 and L2, neither the author nor the second annotator above, worked independently. A box that encloses everything visible of an object counts as one object, and a box with only background, ground, wall or sky has no clear object. Each pool's valid share is corrected for the annotator's accuracy on the controls (Rogan--Gladen) and weighted by group size over all 12{,}139 failures; both annotators are accurate on the controls (sensitivity 0.96 for both, specificity 0.94 and 0.98).

The annotator first recruited as L1 marked seven of the eight test boxes the model labelled sky, wall or field as one object, against the written instruction (corrected estimate 83\%), and we replaced that annotator with a new one; this replacement was decided after seeing that result, and the first annotator's labels are not used. The new L1 marked seven of the eight as no object and one (B156, a sign) as one object; L2 marked all eight as no object.

\begin{center}\footnotesize\setlength{\tabcolsep}{3pt}
\begin{tabular}{p{0.200\linewidth}p{0.370\linewidth}p{0.370\linewidth}}
\toprule
Share of all failures judged valid & L1 & L2 \\
\midrule
uncorrected & 74.6\% & 64.0\% \\
corrected for control accuracy & 77.2\% [67.6, 85.1] & 66.2\% [56.4, 76.2] \\
\bottomrule
\end{tabular}
\end{center}

\emph{Loose boxes count as not valid; 95\% bootstrap intervals over items.}

On the same 100 test boxes and with stuff regions counted not valid, as the instruction says, the author's label-visible labels give 62.5\% (uncorrected), against 74.6\% for L1 and 64.0\% for L2; with stuff regions counted valid they give 71.6\%. Hiding the label therefore did not lower the valid share. L1 treats stuff regions as the instruction says for seven of the eight stuff boxes and L2 for all eight. The annotators agree on 92.0\% of the 200 items ($\kappa$ over three categories 0.82) and on 89\% of the 100 test boxes ($\kappa$ 0.57, PABAK 0.78; $\kappa$ is lower because almost all test boxes are valid), and differ on 11 of the 100 test boxes. Counting a box as valid only if both say so gives 63.5\% of all failures, and if either does 75.1\% (uncorrected). The pre-registration does not say how to combine two annotators, and it contains two rules that point in different directions: the conclusion stands if the label-free and earlier intervals overlap; if the corrected lower bound is below 60\% the claim is lowered and the earlier estimate flagged as an upper bound. Both intervals contain 73\%, and L2's lower bound is 56.4\%. We apply neither rule as a switch and state this as our interpretation: the valid share of the failures is about two thirds to just over three quarters (66\% and 77\%), the like-for-like label-visible figure is 63\%, and 73\% lies inside the range and is not a bound on it. The two intervals together span 56--85\%, and the difference between annotators is larger than the correction. As a check that annotators were not accepting any object, two text-only judges rated whether the word an annotator wrote for a valid box names the same kind of object as the model's label: 79--93\% of pairs depending on annotator and judge, with the judges disagreeing on 8 to 10 pairs per annotator.

\textbf{Recall.} Against COCO's own boxes the pipeline finds 14{,}250 of 36{,}781 ground-truth boxes at IoU 0.5 (recall 38.7\%), so class-agnostic AP at IoU 0.5 cannot exceed 0.387 on COCO's ground truth, where we measure 0.276. Low CAAP therefore has two sources, precision lost to detections of unannotated objects and COCO objects the pipeline does not find; we did not separate their contributions to CAAP.

\subsection{Robustness: a different detector and a different LLM}

We reran the pipeline with each component swapped on the full 5{,}000 images (Grounding DINO for the detector; Qwen2.5-VL 7B for the LLM, with labels regenerated because the original outputs were lost).

\begin{center}\footnotesize\setlength{\tabcolsep}{3pt}
\begin{tabular}{p{0.200\linewidth}p{0.240\linewidth}p{0.240\linewidth}p{0.240\linewidth}}
\toprule
Metric & main (YOLO-World + Gemma-3) & Grounding DINO & Qwen2.5-VL 7B \\
\midrule
native hit rate & 80.9\% & 81.1\% & 80.9\% \\
out-of-vocabulary hit rate & 31.0\% & 25.8\% & 28.3\% \\
native--OOV gap & 49.9 pp & 55.3 pp & 52.6 pp \\
corpus CAAP & 0.207 & 0.229 & 0.211 \\
\bottomrule
\end{tabular}
\end{center}

The vocabulary effect reproduces across two detectors and two LLMs, and under a paraphrased prompt and an ``everyday names'' prompt (gaps 51.8 and 54.8 pp). Two repeats of the main configuration move CAAP by at most 0.003 and SNAP by at most 0.006, far less than the gaps interpreted here. The audits were run on the main configuration only.

\subsection{Checks against other annotation sources}

We re-scored the same detections against two other annotation sources without new inference. LVIS (federated, covering 4{,}809 of our images) matches 4{,}718 of the 11{,}913 COCO failures in those images (39.6\%) at IoU $\geq$ 0.5 (42.4\% of the no-overlap failures, 35.6\% of those overlapping only a differently-labelled box and 48.2\% of the loose boxes); because LVIS is incomplete by design, this is a lower bound (LVIS itself matches only 56.6\% of the detections COCO scores as hits, so with the same completeness the figure would be about 70\%, an assumption we did not test). COCO-ReM, which adds 7{,}217 instances of the same 80 classes, matches only 766 of the 12{,}139 failures (6.3\%), so missing instances of COCO's 80 classes explain only a small part of the failures; most are objects outside them. Scoring the detections class-agnostically against either source does not raise CAAP (COCO-ReM 0.204, LVIS 0.110, against 0.207), because the added boxes include many objects the pipeline does not find.

\section{Discussion}

Vocabulary commonness alone suggests that uncommon vocabulary hurts localization; \S{}3.3 shows that the degradation concentrates where the LLM's wording points the detector at a different object than the one COCO annotated. The LLM, unconstrained by COCO's ontology, names whatever it perceives, including objects and parts outside it. When it names something outside, the audits show that the detector usually localizes that object, but no ground-truth box exists to score against and the evaluation records a failure. For these detections the mismatch is between the evaluation's assumptions and the system's behaviour, not a deficit in visual grounding. A low CAAP is therefore partly evidence that the benchmark's vocabulary is too narrow to score a system not restricted to it, and probably of similar pipelines; B-FOR's independent top-1 audit points the same way. The CAAP--SNAP gap itself is a separate matter, much of which is produced by the encoder (\S{}3.1).

Our data do not test any change to the LLM, detector or prompt, so we make no recommendation about system design. The supported implication is for evaluation: a CAAP-style score on COCO mixes genuine localization errors with detections of objects the benchmark does not label, and the two should be separated, for example by auditing a sample of failures, before the score ranks systems or flags hallucination.

\section{Limitations}

\begin{itemize}\setlength\itemsep{2pt}
\item \textbf{Scope.} All conclusions rest on COCO-Val and its 80-category scheme; we did not evaluate against an exhaustively annotated set. The audits used one configuration (Gemma-3 + YOLO-World), and the other configurations were run once.
\item \textbf{Label-visible audits.} One annotator is an author who wrote the rules and knew the hypothesis, the second is anonymous, and both saw the label, so anchoring is not excluded. The audits give detection-level precision, not CAAP. The 73\% valid share and the 14.3\% loose-box share (a name-matching count, not manually verified) are not on one scale and should not be added.
\item \textbf{Label-free audit.} There are 100 test boxes, so estimates are wide; the two annotators differ by 11 points, one lower bound crosses the pre-registered 60\% threshold, and the pre-registered rules conflict, so the headline range is an interpretation. An annotator who did not follow the written instruction for stuff regions was replaced after their result was seen. The correction assumes control-box error rates hold for the test boxes. Only two annotators contributed.
\item \textbf{Metrics and judges.} Our evaluation is an independent implementation, compared with LAOD's rewrite only through its documented thresholds. The judge classification is partly circular, its text-only variant has no human reference, and the naming check rests on two LLM judges that disagree on some pairs.
\item \textbf{Recall.} Recall against COCO's boxes is 38.7\%; how much CAAP itself would change was not measured.
\end{itemize}

\section{Conclusion}

LAOD reports a gap between localization and naming accuracy, measured with CLIP text similarity, without explaining it. Much of the gap is produced by CLIP at low similarity thresholds: with shuffled labels the CLIP score is 0.33, and with Sentence-BERT SNAP is 0.22 against a CAAP of 0.21. The failures that CAAP counts track vocabulary more than size, and only 14.3\% are loose boxes on the named object, whereas an estimated 73\% (label visible; 63\% with stuff regions not counted) are valid detections of objects COCO does not annotate. A label-free audit by two other annotators, who do not count stuff regions, gives 77\% and 66\%, and one lower bound is below our pre-registered 60\% threshold, so we state the share as about two thirds to just over three quarters; a re-score against LVIS, which does not rely on annotators, gives a lower bound of 39.6\%. The vocabulary effect reproduces with a different detector and LLM; the audits were run on one configuration. For grounded vision-language systems intended for open-world deployment, reported localization failures need to be separated into genuine errors and annotation artifacts before they are used to rank models or to detect hallucination.

\section{Figures}

Figures are numbered by first appearance in the text.

\begin{itemize}\setlength\itemsep{2pt}
\item \textbf{Fig. 1}. Hit rate by detection-box size, split by vocabulary category.
\item \textbf{Fig. 2}. Hit rate by vocabulary category (native, plural of native, out-of-vocabulary). Out-of-vocabulary labels fall to 31.0\% from 80.9\%.
\item \textbf{Fig. 3}. The 12{,}139 IoU-0.5 failures by group; the audits split the first two groups into valid-but-unannotated, ambiguous and wrong, extrapolated from n = 200 and n = 96, so the counts in the first two bars are estimates.
\item \textbf{Fig. 4}. Hit rate of same-object versus different-object labels under three LLM judges, over detections that overlap a ground-truth box. The dashed line is the native rate over all detections (80.9\%; 82.7\% on the bars' denominator).
\item \textbf{Fig. 5}. Manual classification of the no-overlap subset (n = 200); the error bar is the Wilson 95\% CI for the valid-but-unannotated share.
\end{itemize}

\section*{Appendix: Code and data}

The code (CAAP, SNAP, the pipeline wrapper and every analysis script), the 5{,}000-image predictions, the per-detection table, the audit forms, all annotator labels, the pre-registration and the judge prompts and raw responses are released at \url{https://github.com/hollis-png/laod-caap-snap-attribution} (folder \texttt{paper\_\allowbreak{}v2/\allowbreak{}} for the data, audits and results). The analysis scripts are labelled E1--E22 in the order they were run, and \texttt{new\_\allowbreak{}experiments/\allowbreak{}REPORT.\allowbreak{}md} describes each one's inputs and numbers.

\bibliographystyle{plainnat}
\bibliography{references}

@article{laod2025,
  title={{LLM}-Guided Agentic Object Detection for Open-World Understanding},
  author={Mumcu, Furkan and Jones, Michael J. and Cherian, Anoop and Yilmaz, Yasin},
  journal={arXiv preprint arXiv:2507.10844},
  year={2025}
}

@inproceedings{liu2024groundingdino,
  title={Grounding {DINO}: Marrying {DINO} with Grounded Pre-Training for Open-Set Object Detection},
  author={Liu, Shilong and Zeng, Zhaoyang and Ren, Tianhe and Li, Feng and Zhang, Hao and Yang, Jie and Jiang, Qing and Li, Chunyuan and Yang, Jianwei and Su, Hang and Zhu, Jun and Zhang, Lei},
  booktitle={Computer Vision -- ECCV 2024},
  series={Lecture Notes in Computer Science},
  pages={38--55},
  year={2024},
  doi={10.1007/978-3-031-72970-6_3}
}

@inproceedings{gupta2019lvis,
  title={{LVIS}: A Dataset for Large Vocabulary Instance Segmentation},
  author={Gupta, Agrim and Doll{\'a}r, Piotr and Girshick, Ross},
  booktitle={Proceedings of the IEEE/CVF Conference on Computer Vision and Pattern Recognition (CVPR)},
  pages={5351--5359},
  year={2019},
  doi={10.1109/CVPR.2019.00550}
}

@inproceedings{singh2024cocorem,
  title={Benchmarking Object Detectors with {COCO}: A New Path Forward},
  author={Singh, Shweta and Yadav, Aayan and Jain, Jitesh and Shi, Humphrey and Johnson, Justin and Desai, Karan},
  booktitle={Computer Vision -- ECCV 2024},
  series={Lecture Notes in Computer Science},
  pages={279--295},
  year={2024},
  doi={10.1007/978-3-031-72784-9_16},
  note={arXiv:2403.18819}
}

@inproceedings{deng2024coconut,
  title={{COCONut}: Modernizing {COCO} Segmentation},
  author={Deng, Xueqing and Yu, Qihang and Wang, Peng and Shen, Xiaohui and Chen, Liang-Chieh},
  booktitle={Proceedings of the IEEE/CVF Conference on Computer Vision and Pattern Recognition (CVPR)},
  pages={21863--21873},
  year={2024},
  doi={10.1109/CVPR52733.2024.02065},
  note={arXiv:2404.08639}
}

@inproceedings{bolya2020tide,
  title={{TIDE}: A General Toolbox for Identifying Object Detection Errors},
  author={Bolya, Daniel and Foley, Sean and Hays, James and Hoffman, Judy},
  booktitle={Computer Vision -- ECCV 2020},
  series={Lecture Notes in Computer Science},
  pages={558--573},
  year={2020},
  doi={10.1007/978-3-030-58580-8_33}
}

@inproceedings{zhang2026noovd,
  title={{NoOVD}: Novel Category Discovery and Embedding for Open-Vocabulary Object Detection},
  author={Zhang, Yupeng and Han, Ruize and Chen, Zhiwei and Feng, Wei and Wan, Liang},
  booktitle={Proceedings of the IEEE/CVF Conference on Computer Vision and Pattern Recognition (CVPR)},
  year={2026},
  note={arXiv:2603.21069}
}

@inproceedings{batool2026bfor,
  title={Background-Free Objectness Learning for Class-Agnostic Detection},
  author={Batool, Dania and Lo Presti, Liliana and La Cascia, Marco and Vella, Filippo},
  booktitle={British Machine Vision Conference (BMVC)},
  year={2026},
  note={arXiv:2608.29232}
}
\end{document}